\documentclass{article}

\usepackage[final,nonatbib]{nips_2017}

\usepackage[utf8]{inputenc} 
\usepackage[T1]{fontenc}    
\usepackage{hyperref}       
\usepackage{url}            
\usepackage{booktabs}       
\usepackage{amsfonts}       
\usepackage{nicefrac}       
\usepackage{microtype}      
\usepackage{amsmath}
\usepackage[numbers]{natbib}
\usepackage{array}

\usepackage{graphicx}

\title{An Efficient ADMM Algorithm for Structural Break Detection in Multivariate Time Series}

\author{
  Alex Tank, Emily B. Fox, Ali Shojaie \\
  University of Washington\\
  \texttt{alextank@uw.edu, ebfox@uw.edu, ashojaie@uw.edu} \\
}

\begin{document}

\maketitle

\begin{abstract}
We present an efficient alternating direction method of multipliers (ADMM) algorithm for segmenting a multivariate non-stationary time series with structural breaks into stationary regions. We draw from recent work where the series is assumed to follow a vector autoregressive model within segments and a convex estimation procedure may be formulated using group fused lasso penalties. Our ADMM approach first splits the convex problem into a global quadratic program and a simple group lasso proximal update. We show that the global problem may be parallelized over rows of the time dependent transition matrices and furthermore that each subproblem may be rewritten in a form identical to the log-likelihood of a Gaussian state space model. Consequently, we develop a Kalman smoothing algorithm to solve the global update in time linear in the length of the series. 
\end{abstract}

\section{Introduction}
In many applied fields, such as neuroscience and economics, it is necessary to segment a non-stationary and multivariate signal into stationary regimes. Many methods have been proposed to accomplish this important challenge. Bayesian approaches \cite{fox:2009} typically define a generative model, like a vector autoregressive model (VAR), for each stationary regime and a switching Markov process to model switches between regimes. More nonparametric methods \cite{ombao:2005, preuss:2013} directly analyze jumps in the spectral density of the process over time.  

Recently, many authors have explored segmentation procedures based on convex optimization \cite{chan:2014, chan:2015,Safikhani:2017}. Similar to Bayesian methods, convex approaches model each regime using an autoregressive model. Furthermore, fused group lasso penalties enforce the constraint that the autoregressive parameters of the process tend to stay constant over time, and only rarely switch to new parameter values.  In practice, segmentation methodologies using fused group lasso penalties have relied on an approximate group least angle regression \cite{yuan:2006} solver for optimization \cite{chan:2014, chan:2015}. While an intuitive and widely used algorithm, group least angle regression does not provide any garauntees for returning the optimal solution for the convex segmentation problem. 

Instead, we develop an efficient alternating direction method of multipliers (ADMM) algorithm \cite{boyd:2011} that directly solves the convex segmentation problem with group fused lasso penalties. Our ADMM approach splits the convex problem into a global quadratic program that may be solved in time linear with the series length and a simple group lasso proximal update.

Both code for the ADMM algorithm and code to reproduce our experiments may be found at \texttt{bitbucket.org/atank/convex\_tar}.

\section{Background}
Let $x_t \in \mathbb{R}^p$ be a $p$-dimensional multivariate time series. We assume that $x_t$ follows a structural break vector autoregressive model. Specifically, let $L$ be the number of break points occurring at times $\left(t_1, \ldots, t_L \right)$. For each $t \in (t_i, t_{i + 1}]$, $x_t$ follows a stationary vector autoregressive model (VAR) of lag order $K$ 
\begin{align}
x_t = \sum_{k = 1}^K A^{ik} x_{t - k} + e_t,
\end{align}
where $\left(A^{ik}, \ldots, A^{iK}\right)$ are the $K$ $p \times p$ matrices of the $i$th VAR process and $e_t \in \mathbb{R}^p$ is mean zero noise. 

Given an observed time series at $N$ time points, $\left( x_1, \ldots, x_N \right)$, the goal of estimation is to segment the series into $\hat{L} + 1$ stationary blocks, where $\hat{L}$ is the estimated number of change points. To do this, estimates of the breakpoints, $\left(\hat{t}_1, \ldots, \hat{t}_{\hat{L}} \right)$, and estimates of the autoregressive VAR parameters, $\hat{A}^{ik}$ for $k \in (1, \ldots, K)$ and $i \in (1, \ldots, \hat{L})$, in each stationary segment must be determined.

\section{Estimation}
We follow previous work and formulate structural break estimation in autoregressive models via a convex optimization problem with fused group lasso penalties \cite{Safikhani:2017,chan:2015,chan:2014}. First, we introduce local autoregressive parameters $A^t = \left(A^{t1}, \ldots, A^{tK} \right)$ active at each time point. We then solve the following penalized least squares optimization problem
\begin{align} \label{csol}
\min_{A^1, \ldots, A^N} \sum_{t = 1}^N||x_t - A^{t} \tilde{x}_{t}||_2^2 + \lambda \sum_{t = 2}^N ||A^t - A^{t + 1}||_F,
\end{align}
where  $\tilde{x_t} = (x_{t - 1}^T, \ldots, x_{t - K}^T)^T$, $||.||_F$ is the Frobenius norm that acts as a group lasso penalty and $\lambda > 0$ is a tuning parameter that controls the number of estimated break points. In this setting, the fused group lasso penalty shrinks the $A^t$ and $A^{t + 1}$ parameter estimates to be identical.  The change point estimates $(t_1, \ldots t_{\hat{L}})$ are those times $t$ whenever $\hat{A}^t \neq \hat{A}^{t + 1}$, where $(\hat{A}^1, \ldots, \hat{A}^N)$ is the solution to Problem (\ref{csol}), and $\hat{L}$ is the number of such time points.

\section{ADMM Algorithm}
To solve Problem (\ref{csol}) exactly we develop an efficient ADMM algorithm that takes advantage of the time series structure. First, we introduce a change of variables parameterization $\theta^1 = A^1$ and $\theta^t = A^t - A^{t + 1}$ for $t > 1$. The reparameterization lets us rewrite Problem (\ref{csol}) as
\begin{align} \label{reparam}
\min_{\theta^1, \ldots, \theta^N} \|Y - X \theta\|_F^2 + \lambda \sum_{t = 2}^N \|\theta^t\|_F,
\end{align}
where $\theta = (\theta^{1}, \ldots, \theta^{N})^T$, $Y = (x_1, \ldots, x_N)^T$ and
\begin{align}
X = \left( \begin{matrix}
\tilde{x}_1^T & 0 & 0 & \ldots & 0 \\
\tilde{x}_2^T & \tilde{x}_2^T & 0 & \ldots & 0 \\
\tilde{x}_3^T & \tilde{x}_3^T & \tilde{x}_3^T & \ldots 0 \\
\vdots & \vdots & \vdots & \ddots & \vdots \\
\tilde{x}_N^T & \tilde{x}_N^T & \tilde{x}_N^T & \ldots & \tilde{x}_N^T
\end{matrix}
\right).
\end{align}

Since Problem (\ref{reparam}) takes the form of a group lasso regression problem, approximate solvers like group least angle regression may be used \cite{chan:2015,chan:2014}. However, we instead develop an efficient ADMM algorithm to solve Problem (\ref{reparam}) exactly. 

As is standard in ADMM, we introduce the parameter $W = (W^{1}, \ldots, W^{N})^T$ and the constraint $W = \theta$ to break apart the least squares term and the group lasso penalty in Problem (\ref{reparam})

\begin{align} \label{constraint}
\min_{W, \theta} \|Y - X \theta\|_F^2 + \lambda \sum_{t = 2}^N \|W^t\|_F.
\end{align}

The augmented Lagrangian for Problem (\ref{constraint}) is given by

\begin{align}
\min_{W, \theta} \|Y - X \theta\|^2_F + \lambda \sum_{t = 1}^N \|W^t\|_F + \frac{\rho}{2} \|\theta - W\|^2_F +  \text{trace}(\Omega^T ( \theta - W)),
\end{align}

where $\Omega \in \mathbb{R}^{p \times p K N}$ are Lagrange multipliers and $\rho > 0$. The scaled ADMM steps for solving the augmented Lagrangian are given by \cite{boyd:2011}
\begin{align}
\theta^{(l + 1)} &= \min_{\theta} \|Y - X \theta\|_F^2 + \frac{\rho}{2}\| \theta - W^{(l)} + \Omega^{(l)}\|_F^2 \label{global} \\
W^{(l + 1)} &= \min_{W} \lambda \sum_{i = 2}^N \|W^i\|_F + \frac{\rho}{2} \| \theta^{(l + 1)} - W + \Omega^{(l)}\|_F^2 \label{local}\\
\Omega^{(l + 1)} &= \Omega^{(l)} + \theta^{(l + 1)} - W^{(l + 1)}.
\end{align}
We present the specific form for solving Problems (\ref{global}) and (\ref{local}) in Sections \ref{global_det} and \ref{local_det} below.

\subsection{Global $\theta$ update} \label{global_det}
Although the global ADMM subproblem given in Eq. (\ref{global}) is a quadratic program with $p^2 K N$ variables, we develop an efficient $\mathcal{O}(N)$ linear time algorithm for its solution. First, we reintroduce the $\theta^t = A^{t + 1} - A^{t}$ parameterization, which gives
\begin{align} \label{rewrite}
\min_{A^1, \ldots, A^N} \sum_{t = 1}^N \|x_t - A^t \tilde{x}_t\|_2^2 + \frac{\rho}{2} \sum_{t = 1}^N \|A^{t + 1} - A^{t}  - W^{t(l)} + \Omega^{t(l)} \|_F^2
\end{align}
Furthermore, Problem (\ref{rewrite}) may be decomposed into $p$ independent problems which may be solved in parallel for each row of $A^t = \left(a^t_1, \ldots, a^t_{p}\right)^T$. The problem for each  $(a^1_j, \ldots, a^N_j)$ is given by

\begin{align} \label{rowa}
\min_{a_j^1, \ldots, a_j^N} \sum_{t = 1}^N \left(x_{tj} - \tilde{x}_t^T a_j^t \right)^2 + \frac{\rho}{2} \sum_{t = 1}^N \|a^{t + 1}_j - a^{t}_j  - W^{t(l)}_j + \Omega^{t(l)}_j \|_2^2. 
\end{align}
Problem (\ref{rowa}) may be solved efficiently by noting that it takes the same form as a canonical smoothing problem for the $\left(a^1_j, \ldots, a^N_j \right)$ in a state space model. Specifically, Problem (\ref{rowa}) is the negative log-likelihood of a Gaussian state space model \cite{commandeur:2007} of the following form:
\begin{align}
a_j^t &=  a_j^{t - 1} + \mu_t + \gamma_t  \\
x_{tj} &= \tilde{x}_t^T a_j^t+ \eta_t
\end{align}
where by convention $a_j^0 = 0$, $\mu_t =  W^{t(l)}_j - \Omega^{t(l)}_j$ is the bias added at each time step, $\gamma_t \sim N(0, \frac{1}{\rho}I_{pK \times pK})$ is the state evolution noise with covariance matrix $\frac{1}{\rho}I_{pK \times pK}$, and $\eta_t \sim N(0,\frac{1}{2})$ is the observation noise with variance $\frac{1}{2}$.

Inference for the maximum likelihood state sequence $\left(a^1_j, \ldots, a^N_j \right)$ in this model may be solved using a \emph{Kalman filtering-smoothing} algorithm. Kalman smoothers compute the expected value of the latent sequence given the observations and due to Gaussianity this expected value is the same as the mode of the log-likelihood. Many such smoothing algorithms exist but here we employ the classical Rauch-Tung-Streibel smoother \cite{commandeur:2007}. In our case, this smoothing algorithm reduces to first computing the following forward filtering steps initialized with $\hat{a}^{1 | 1}_j = \mu_1$ and $\Sigma^{1 | 1} = \rho^{-1} I_{pK \times pK}$ then recursively computing for $t > 1$:
\begin{align}
\hat{a}^{t | (t - 1)}_j &= \hat{a}^{(t - 1) | (t - 1)}_j + \mu_t \nonumber \\
\Sigma^{t|(t-1)} &= \Sigma^{(t-1)|(t-1)} + \rho^{-1} I_{pK \times pK} \nonumber \\
K^{t} &= \frac{\Sigma^{t|(t-1)} \tilde{x}_t}{\frac{1}{2} + \tilde{x}_t^T \Sigma^{t|(t-1)} \tilde{x}_t} \nonumber \\
\hat{a}^{t | t}_j &= \hat{a}^{t | (t - 1)}_j + K^t \left(x_{tj} - \tilde{x}_t^T \hat{a}^{t | (t - 1)}_j \right) \nonumber \\
\Sigma^{t| t} &= (I - K^t \tilde{x}_t^T) \Sigma^{t | (t - 1)}, \nonumber
\end{align}
followed by the backward smoothing steps initialized with $\hat{a}^N = \hat{a}^{N|N}$ and $\Sigma^N = \Sigma^{N|N}$:
\begin{align}
\hat{a}^t_j &= \hat{a}^{t | t}_j + C^t \left( \hat{a}^{t + 1} - \hat{a}^{t + 1|t} \right) \nonumber \,\,\,\,\,\,\,\,\,\, \Sigma^t = \Sigma^{t|t} + C^t \left( \Sigma^{t + 1} - \Sigma^{t + 1 | t} \right) C^{tT}, \nonumber
\end{align}
where $C^t = \Sigma^{t|t} \left(\Sigma^{(t + 1)|t} \right)^{-1}$ and $(\hat{a}^1_j, \ldots, \hat{a}_j^N)$ is the optimal solution to Problem (\ref{rowa}). Since both forward and backward passes each are linear in $N$, the full smoothing computation to solve Problem (\ref{rowa}) is linear in $N$.
\subsection{W update} \label{local_det}
The $W$ update in Problem (\ref{local}) is given separately for each $W^t$. Specifically, it is given by the proximal operator for the group lasso penalty, which is a group soft threshold step
\begin{align}
W^{t(l + 1)} = \begin{cases}
0 \,\,\,\,\,\, &\text{if } \|\theta^{t (l+ 1)} + \Omega^{t(l)}\|_F < \frac{\lambda}{\rho} \\
\left(1 - \frac{\lambda}{\rho \|\theta^{t (l+ 1)} + \Omega^{t(l)}\|_F } \right) \left( \theta^{t (l+ 1)} + \Omega^{t(l)} \right) & \text{otherwise.}
\end{cases}
\end{align}

\section{Simulation}
\begin{figure} \label{breakpoints}
\centering
\includegraphics[width=.5\textwidth]{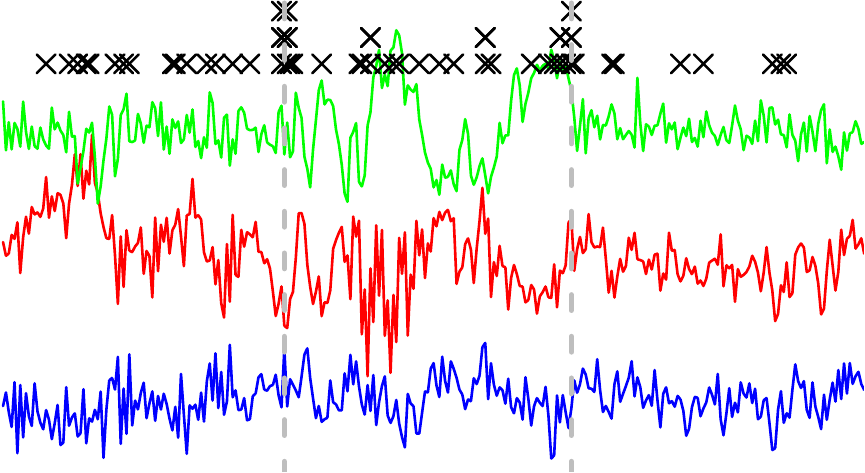}
\caption{Three of ten series series from a $p = 10$ $N = 300$ series with estimated change points. Each row of $\times$s indicate detected change points for a different $\lambda \in (1,3,5)$. True change point times shown in dotted grey.}
\end{figure}

To test our algorithm we detect breakpoints on a $p = 10$ series with length $N = 300$ observations. We randomly generate a series with two structural break points at times $t \in (100,200)$ for a total of three stationary regions each generated by a different VAR(1) process. We run our algorithm for three lambda settings, $\lambda \in (1,3,5)$, and a breakpoint is detected if $||\theta^t||_F > .005$. The estimated break points are shown in Figure \ref{breakpoints}. Overall, the $\lambda = 5$ case acurately detects the breakpoint times.
\section{Discussion and Future Work}

The global step of our ADMM algorithm solves $p$ independent smoothing problems. While the Kalman filter we utilize has runtime linear in $N$, each recursive step during the backward smoothing phase requires the inverse of a $p K \times p K$ matrix. While this computation is viable for moderate sized $p$, future work aims to explore alternate Gaussian state space smoothers that scale better with $p$ for high dimensional applications. 

\paragraph{Acknowledgments}
AT and EF acknowledge the support of ONR Grant N00014-15-1-2380,
NSF CAREER Award IIS-1350133. AS acknowledges the
support from NSF grants DMS-1161565 and DMS-1561814 and NIH grants 1K01HL124050-01 and
1R01GM114029-01.

\bibliographystyle{plain}
\bibliography{nips_ts}

\begin{thebibliography}{1}

\bibitem{boyd:2011}
Stephen Boyd, Neal Parikh, Eric Chu, Borja Peleato, and Jonathan Eckstein.
\newblock Distributed optimization and statistical learning via the alternating
  direction method of multipliers.
\newblock {\em Foundations and Trends{\textregistered} in Machine Learning},
  3(1):1--122, 2011.

\bibitem{chan:2014}
Ngai~Hang Chan, Chun~Yip Yau, and Rong-Mao Zhang.
\newblock Group lasso for structural break time series.
\newblock {\em Journal of the American Statistical Association},
  109(506):590--599, 2014.

\bibitem{chan:2015}
Ngai~Hang Chan, Chun~Yip Yau, and Rong-Mao Zhang.
\newblock Lasso estimation of threshold autoregressive models.
\newblock {\em Journal of Econometrics}, 189(2):285--296, 2015.

\bibitem{commandeur:2007}
Jacques~JF Commandeur and Siem~Jan Koopman.
\newblock {\em An introduction to state space time series analysis}.
\newblock Oxford University Press, 2007.

\bibitem{fox:2009}
Emily Fox, Erik~B Sudderth, Michael~I Jordan, and Alan~S Willsky.
\newblock Nonparametric bayesian learning of switching linear dynamical
  systems.
\newblock In {\em Advances in Neural Information Processing Systems}, pages
  457--464, 2009.

\bibitem{ombao:2005}
Hernando Ombao, Rainer Von~Sachs, and Wensheng Guo.
\newblock Slex analysis of multivariate nonstationary time series.
\newblock {\em Journal of the American Statistical Association},
  100(470):519--531, 2005.

\bibitem{preuss:2013}
P.~{Preu{\ss}}, R.~{Puchstein}, and H.~{Dette}.
\newblock {Detection of multiple structural breaks in multivariate time
  series}.
\newblock {\em ArXiv e-prints}, September 2013.

\bibitem{Safikhani:2017}
A.~{Safikhani} and A.~{Shojaie}.
\newblock {Structural Break Detection in High-Dimensional Non-Stationary VAR
  models}.
\newblock {\em ArXiv e-prints}, August 2017.

\bibitem{yuan:2006}
Ming Yuan and Yi~Lin.
\newblock Model selection and estimation in regression with grouped variables.
\newblock {\em Journal of the Royal Statistical Society: Series B (Statistical
  Methodology)}, 68(1):49--67, 2006.

\end{thebibliography}

\end{document}